\documentclass[letterpaper, 10 pt, conference]{ieeeconf}  % Comment this line out if you need a4paper

\IEEEoverridecommandlockouts                              % This command is only needed if 
\usepackage{xcolor}
\usepackage{float}
\usepackage{graphicx}
\usepackage{booktabs}
\usepackage{multirow}
\usepackage{amsmath}
\usepackage{makecell}
\usepackage{enumerate}
\usepackage{bbm}
\usepackage{dsfont}

\usepackage{array}
\newcolumntype{C}[1]{>{\centering\arraybackslash}p{#1}}

\title{\LARGE \bf
IRef-VLA: A Benchmark for Interactive Referential Grounding with Imperfect Language in 3D Scenes
}

\author{
Haochen Zhang$^{*,1}$ \and
Nader Zantout$^{*,1}$ \and
Pujith Kachana$^{1}$ \and
Ji Zhang$^{1}$ \and
Wenshan Wang$^{1}$%
\thanks{$^{*}$ Denotes equal contribution.}
\thanks{$^{1}$The authors are with Carnegie Mellon University, Robotics Institute, Pittsburgh, PA.
{\tt\small\{haochen4, nzantout, pkachana, zhangji, wenshanw\}@andrew.cmu.edu}}
}

\begin{document}

\maketitle
\thispagestyle{empty}
\pagestyle{empty}

%%%%%%%%%%%%%%%%%%%%%%%%%%%%%%%%%%%%%%%%%%%%%%%%%%%%%%%%%%%%%%%%%%%%%%%%%%%%%%%%
\begin{abstract}
With the recent rise of large language models, vision-language models, and other general foundation models, there is growing potential for multimodal, multi-task robotics that can operate in diverse environments given natural language input. One such application is indoor navigation using natural language instructions. However, despite recent progress, this problem remains challenging due to the 3D spatial reasoning and semantic understanding required. Additionally, the language used may be imperfect or misaligned with the scene, further complicating the task. To address this challenge, we curate a benchmark dataset, IRef-VLA, for Interactive Referential Vision and Language-guided Action in 3D Scenes with imperfect references. IRef-VLA is the largest real-world dataset for the referential grounding task, consisting of over 11.5K scanned 3D rooms from existing datasets, 7.6M heuristically generated semantic relations, and 4.7M referential statements. Our dataset also contains semantic object and room annotations, scene graphs, navigable free space annotations, and is augmented with statements where the language has imperfections or ambiguities. We verify the generalizability of our dataset by evaluating with state-of-the-art models to obtain a performance baseline and also develop a graph-search baseline to demonstrate the performance bound and generation of alternatives using scene-graph knowledge.
With this benchmark, we aim to provide a resource for 3D scene understanding that aids the development of robust, interactive navigation systems. The dataset and all source code is publicly released\footnote{https://github.com/HaochenZ11/IRef-VLA}. 
\end{abstract}

%%%%%%%%%%%%%%%%%%%%%%%%%%%%%%%%%%%%%%%%%%%%%%%%%%%%%%%%%%%%%%%%%%%%%%%%%%%%%%%%
\section{Introduction}

% 2D to 3D, human language and reasoning
Methods combining vision and language have been evolving rapidly with the advent of both Large Language Models (LLMs) \cite{achiam2023gpt, touvron2023llama, team2023gemini} and Vision-Language Models (VLMs) \cite{radford2021learning, ramesh2021zero, liu2024visual} pre-trained on internet-scale data, tackling various 2D tasks such as Visual Question Answering (VQA) \cite{antol2015vqa}, image retrieval \cite{karpathy2014deep}, and image captioning \cite{radford2021learning}. As we progress towards generalizable embodied intelligence, there is a need for methods that are capable of reasoning in 3D-space and interacting with humans. Particularly, being able to understand natural language and ground that language to the physical world are key skills for interactive robots. For example, humans are able to precisely refer to objects in a 3D scene, often using the statement of ``least effort" \cite{zipf2016human} and making use of spatial relationships. An agent that can 1) similarly solve such a problem, 2) handle imperfect or ambiguous language, and 3) interact with humans to achieve the intended goal would be valuable in robotics fields such as indoor-navigation with applications as in-home assistants.

\begin{figure}[H]
\centering
\includegraphics[width=0.4\textwidth]{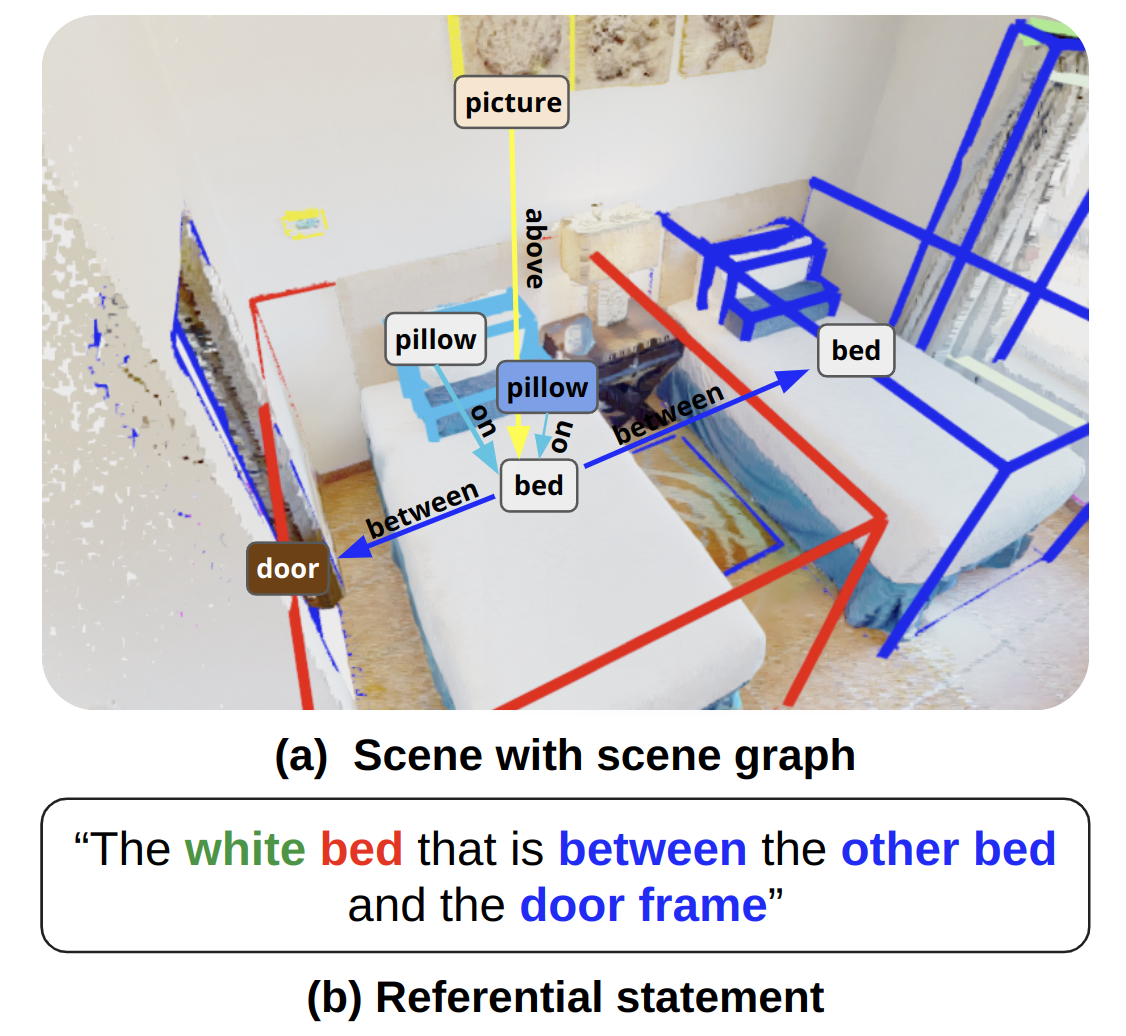}
\caption{Sample region from the dataset visualized with (a) a scene graph and (b) a corresponding referential statement}. 
\label{fig:data_sample}
\end{figure}

%Real-world deployment - robustness, interaction, mistakes
The pursuit of such agents that can identify and understand 3D scenes, consolidate visual input with language semantics, and display robust performance for real-world deployment, however, presents various challenges. First, the scene can have hundreds of objects, contain objects belonging to fine-grained classes, and have many similar objects \cite{ramakrishnan2021habitat}. Second, human referential language often involves spatial reasoning, implicit and explicit affordances, open-vocabulary language, and may even be incorrect or refer to something that does not exist, e.g. \textit{``the remote on the table"} when the remote is actually on the sofa. Third, the scale of available vision-language data in the 3D space pales in comparison to the amount of 2D data, which was crucial to the success of 2D vision-language learning methods \cite{krishna2017visual, changpinyo2021conceptual}. Despite impressive recent advancements with foundation models, such problems remain difficult when applied to robotics as current methods fail to offer the accuracy and robustness needed for real-world deployment \cite{huang2022multi}.

To advance the path towards more intelligent interaction in natural language navigation, we propose the IRef-VLA dataset as a benchmark for both the referential object-grounding task, and a novel extension of this task we call \textbf{referential grounding with imperfect references}. First, we provide the largest real-world dataset based on 3D scenes from a diverse set of existing indoor scans. Our dataset includes 1) segmented scene point clouds to enable learning directly from 3D visual information, 2) object-level attributes, semantic class labels, and affordances, 3) dense scene graphs with spatial relations as structural guidance, 4) heuristically-generated referential statements improving upon previous datasets, 5) traversable free space annotations allowing for references to areas and spaces, and 6) augmented "imperfect" referential statements to benchmark grounding with imperfect language. In particular, the inclusion of scene graphs, free space annotations, and imperfect statements distinguishes our dataset from previous ones. Second, with the inclusion of imperfect language, we define the extended task of referential grounding with imperfect references, testing a model's capability to a) detect when a specific referenced object does not exist in the scene, and b) prompt interaction by generating valid alternative suggestions. A sample from our dataset is shown in Fig. \ref{fig:data_sample}. 

To validate our dataset, we train two state-of-the-art (SOTA) supervised referential grounding models on our dataset and demonstrate generalizability to test benchmarks. We also implement our own graph-search method that first determines whether an object is in the scene, then suggests alternatives if needed. We compare performance to an augmented SOTA model that classifies object existence. We release our dataset, source code for generation and baselines, and a dataset visualization tool publicly.

\section{Related Work}
\subsection*{Object Referential Datasets}
% Referit3d, ScanRefer, SceneVerse
The referential object-grounding task has been defined and explored in 3D datasets such as ReferIt3D \cite{achlioptas2020referit3d}, ScanRefer \cite{chen2020scanrefer}, and SceneVerse \cite{jia2024sceneverse}. While ReferIt3D and ScanRefer establish a benchmark for referential grounding on one set of scenes providing both synthetically generated (Sr3D) and human-uttered (Nr3D) statements \cite{achlioptas2020referit3d}, ambiguities exist in the synthetic statements and the human utterances can be subjective and unintuitive, e.g. using the word ``\textit{comfy}" or using clock bearings to describe objects. SceneVerse scales the data up by curating a much larger dataset and generating statements synthetically, then using an LLM for rephrasing, though both templated and LLM-rephrased statements are often unnatural and lack explicit references to attributes like size, color, and shape which humans often use for object reference. As a result, models trained on SceneVerse still performed poorly on the Nr3D benchmark \cite{jia2024sceneverse}. More recently, SpatialRGPT \cite{cheng2024spatialrgptgroundedspatialreasoning} proposes a data pipeline to further push the scale of 3D-grounded referential statements using existing large VLMs, although it is limited to 2D images and does not directly contain 3D data.

\subsection*{Semantic Scene Graph Datasets}
% 3DSSG, Hydra, HOV-SG, ConceptGraphs
Generating scene graphs from 3D scenes has also been explored in 3DSSG \cite{wald2020learning}, Hydra \cite{hughes2022hydra}, HOV-SG \cite{werby2024hierarchical}, and ConceptGraphs \cite{gu2023conceptgraphs}. 3DSSG focuses on predicting scene graphs automatically, resulting in generated graphs that can miss relations or generate redundant ones, which requires more processing to disambiguate objects given their relations. In Hydra, a system is developed to build 3D scene graphs in real-time but does not include explicit language-grounding. While HOV-SG and ConceptGraphs both build open-vocabulary scene graphs, they are designed for referring to an object mainly using region references rather than fine-grained inter-object relations.

% \vspace{-2em}
\subsection*{Instruction-Following Datasets}
Multiple works have also explored language-guided navigation through instruction-following statements, often specifying a series of steps to move between regions in a large scene. Common datasets including Room Across Room \cite{ku2020room} and Room-2-Room \cite{mattersim} focus on generating distinct steps to navigate between rooms but do not explicitly focus on decomposing the task into disambiguating objects explicitly through spatial relations, making it difficult to be robust to scene changes or imperfect language.

% \vspace{-3em}
\subsection*{Referential Object Grounding}
A number of papers have explored the task of learning referential object grounding, mainly on either the ReferIt3D benchmark or the ScanRefer task. These include BUTD-DETR \cite{jain2022bottom}, MVT \cite{huang2022multi}, ViL3DRel \cite{chen2020scanrefer}, 3D-VisTA \cite{zhu20233d}, and GPS trained on SceneVerse \cite{jia2024sceneverse}. Despite massively upscaling data, GPS still only achieves an accuracy of 64.9\% on Nr3D \cite{jia2024sceneverse} and has low zero-shot generalization capabilities to new scenes and language. These models are also incapable of handling ambiguities in language input, and with the exception of GPS, cannot handle open-vocabulary input, making them unideal for real-world deployment.

\subsection*{Language Interaction in Embodied Agents}
Some works \cite{pramanick2022talk, shridhar2018interactive, zhang2021invigorate}, have explored the task of interactive visual grounding and ambiguity resolution, however, the formulation is either limited to simple input statements in 2D images, or the evaluation of ambiguous statements is limited to small amounts of human-annotated data on few scenes due to the cost and lack of such data. Other work in embodied, interactive agents has focused on multi-turn natural language dialogue. The TEACh benchmark \cite{padmakumar2021teachtaskdrivenembodiedagents} offers a human-generated dataset of task-driven dialogues for language grounding, dialogue understanding and task reasoning. 
\cite{sharma2022correctingrobotplansnatural} demonstrates the benefits of language feedback for improving real-world robotics tasks, although it is limited to one-way communication as the agent cannot pose questions to the human user. Extending this to navigation, \cite{shah2022lmnavroboticnavigationlarge} presents an instruction-following navigation system that uses large pretrained models, showing the effectiveness of large-scale data for language-guided navigation tasks. Building on these advancements, we aim to enhance interactive navigation in 3D scenes by improving the scale and quality of 3D language data with a focus on language scenarios that prompt further interaction for spatial raesoning.

\section{Task Formulation}

\subsection{Referential Grounding with Imperfect References Task}
We define the task of referential grounding with imperfect references as an augmented version of referential object-grounding which involves identifying objects without assuming a perfect match between references and scene objects. For a given statement, a referred object is only returned if it exists, otherwise the expected response is a) an explicit indication that the object was not found, and b) a suggested alternative object. We see this as an initial step in interactive referential grounding, which can facilitate navigation by clarifying uncertainties about the intended goal.

We differentiate this task from the original referential object-grounding task in ReferIt3D \cite{achlioptas2020referit3d} and embodied tasks such as ObjectNav \cite{batra2020objectnav}, ObjectGoal \cite{anderson2018evaluation}, and AreaGoal \cite{anderson2018evaluation}. Compared to the standalone object referential grounding task in benchmarks such as ReferIt3D \cite{achlioptas2020referit3d} and ScanRefer \cite{chen2020scanrefer}, our task aims to be a stepping stone towards multi-turn interactive navigation. Instead of assuming the referred object is always present and that only a single retrieval attempt is allowed, our task accommodates imperfect references and allows for multi-turn interactions.

In contrast to existing embodied navigation tasks like Object-Goal Navigation (ObjectNav) \cite{batra2020objectnav}, which evaluate how an agent navigates to a goal, our task focuses on the nuances of 3D language grounding, independent of agent actions or planning. Conversely, the ObjectGoal and AreaGoal tasks are navigation-focused and the statements involve simple references like "find chair," while our task addresses the challenge of grounding complex spatial relations and detailed classes from more complex statements.

In general, we find that current formulations for related tasks are limited by their reliance on simple references, assumptions of reference correctness, and single-shot design. These constraints are unrealistic given the imperfect and dynamic nature of real-world scenes and instructions, highlighting the need for tasks focused on robust grounding in such scenarios.

\subsection{Metrics}
For the grounding and search subtask, we use binary classification metrics—true positive (TP), false positive (FP), true negative (TN), and false negative (FN)—to assess how well the model can identify object existence based on a referential statement.

To quantitatively assess the quality of retrieved object alternatives, we use a heuristic scoring system. We calculate a similarity score \(score_{sim}\) based on how well each suggestion matches aspects of the referential statement, such as object classes, attributes, and spatial relations. Aspects are weighted by importance, with object class and relation given higher weights as these are closer aligned to human intent compared to attributes. The score is then normalized by the maximum possible match score. For a given imperfect referential statement \(S\) with \(n\) total aspects \(A(S) = \{a_1, ... a_n\}\) ordered by class then attributes and \(n\) varying between statements, a selected alternative \(S'\) with \(m\) total aspects \(A(S') = \{a'_1, ... a'_m\}\), and \(\lambda_i\) as the weight on the \(i\)-th aspect of \(S\):

\begin{equation}
score_{sim} = \frac{\Sigma_{i}^{n}\lambda_i * \mathds{1}{\{a_i \in A(S')\}}}{\Sigma_{i}^{n}\lambda_i}
\end{equation}

These heuristics provide a preliminary comparison metric for retrieved alternatives. However, such suggestions depend heavily on original user intent and preferences. Thus, human-labeled scores may better quantify quality, though this approach may be limited in scale.

\section{Dataset Creation}
% \vspace{-2em}
\subsection{Overview}
To advance robust interactive navigation agents, we introduce IRef-VLA, a synthetically-generated public benchmark dataset. It combines 3D scans from five real-world datasets: ScanNet \cite{dai2017scannet}, Matterport3D \cite{Matterport3D}, Habitat-Matterport 3D (HM3D) \cite{ramakrishnan2021habitat}, 3RScan \cite{Wald2019RIO}, and ARKitScenes \cite{baruch2021arkitscenes}, as well as Unity-generated scenes. Fig.  \ref{fig:data_sources} shows the distribution of regions from each source. Each scene includes:
\begin{figure}
\centering
\includegraphics[width=0.36\textwidth]{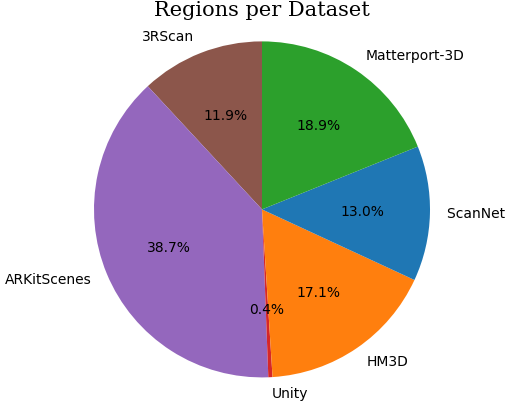}
\setlength{\belowcaptionskip}{-12pt}
\caption{Breakdown of regions from each data source}
\label{fig:data_sources}
\end{figure}

\begin{itemize}
\item Scene point cloud
\item List of objects with semantic class labels, bounding box, and color(s)
\item List of traversable free spaces
\item List of regions with semantic labels and bounding boxes
\item Scene graph of spatial relations split by room
\item Language statements with ground-truth annotation
\end{itemize}
% \vspace{-2em}

\begin{figure}
\centering
\includegraphics[width=0.42\textwidth, trim={0 0cm 0 0cm}]{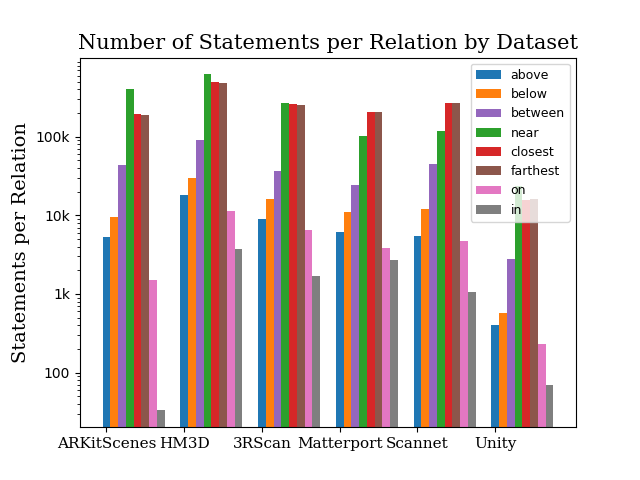}
\caption{Number of statements per relation type from each dataset processed}
\label{fig:relation_statements}
\end{figure}

\begin{figure*}[t!]
\centering
\includegraphics[width=0.8\textwidth, trim={0 0cm 0cm 0cm},clip]{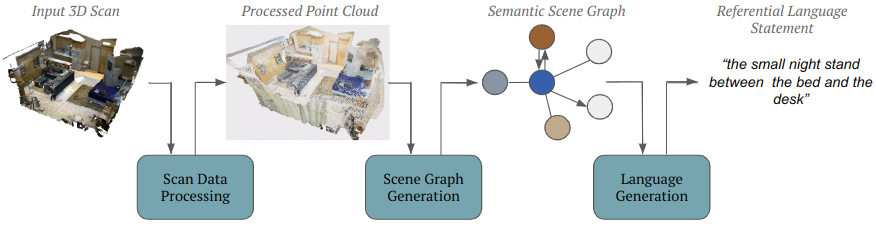}
\caption{Data processing pipeline consisting of: 3D Scan Processing, Scene Graph Generation, and Language Generation}
\label{fig:data_processing}
\end{figure*}

Key features of our dataset are large-scale scene graphs enabling identification of similar objects, traversable free space annotations, and imperfect language statements. In total, our dataset comprises 7,635 scenes with over 11.5K regions, 286K objects across 477 classes, 7.6 million inter-object spatial relations, and 4.7 million referential statements. Fig. \ref{fig:relation_statements} shows the spatial relations per dataset, with further details on the data curation process in Fig. \ref{fig:data_processing}.

\subsection{3D Scan Processing}
To generate point cloud files, we used scene-level point clouds from PLY files for ARKitScenes, Matterport3D, and ScanNet. For HM3D, Unity, and 3RScan, point clouds were uniformly sampled from meshes, with colors derived from textures. Regions and objects were identified using semantic information from the original meshes. ARKitScenes, 3RScan, and ScanNet have single-room scenes, while Matterport3D and HM3D provide region segmentations, and Unity scenes are custom-segmented. Each object is labeled with an open-vocabulary class name, mapped to NYU40 \cite{gupta2013perceptual} and NYUv2 \cite{silberman2012indoor} schemas with the provided mappings\footnote{For the Unity scenes, the ground-truth semantic labels were cleaned then manually mapped to the class schemas by five data annotators. A validation round was done to standardize the labels.}. The dominant three colors were obtained for each object based on the point cloud and a color clustering algorithm.

To provide extra navigation targets, each scan was also processed to generate the horizontally traversable free space. Separate traversable regions in a room are combined into sub-regions, for which spatial relations with other objects in the scene are generated to create unambiguous references to these spaces (e.g. ``the space near the table").

\subsection{Scene Graph Generation}
Eight different types of semantic spatial relations were heuristically calculated based on the yawed object bounding boxes to generate a scene graph of relations. Relations are generated exhaustively for every pair or triple of objects within a region, then filtered based on the semantic classes involved. 
Table \ref{tab:relations} defines the types of spatial relations used.
\begin{table}[t!]
\caption{Summary of semantic relationship types in IRef-VLA}
\label{tab:relations}
\vspace{-2em}
\begin{center}
\resizebox{\linewidth}{!}{
  \begin{tabular}{|l|p{3.5cm}|p{2.5cm}|l|}
     \hline
     \textbf{Relation} & \textbf{Definition} & \textbf{Synonyms} & \textbf{Properties} \\ 
     \hline
     Above & Target is above the anchor & Over &  \\
     \hline
     Below & Target is below the anchor & Under, Beneath, Underneath &  \\
     \hline
     Closest & Target is the closest object of a certain class to the anchor & Nearest & Inter-class \\
     \hline
     Farthest & Target is the farthest object from a certain class to the anchor & Most distant, Farthest away & Inter-class \\
     \hline
     Between & Target is between two anchors & In the middle of, In-between & Ternary \\
     \hline
     Near & Target is within a threshold distance of the anchor & Next to, Close to, Adjacent to, Beside & Symmetric \\
     \hline
     In & Target is inside the anchor & Inside, Within &  \\
     \hline
     On & Target is above and in contact with the anchor in the Z-axis & On top of &  Contact \\
     \hline
  \end{tabular}}
\end{center}
% \vspace{-3em}
\end{table}

\subsection{Language Generation}
\begin{figure}
    \centering
    \includegraphics[width=\linewidth]{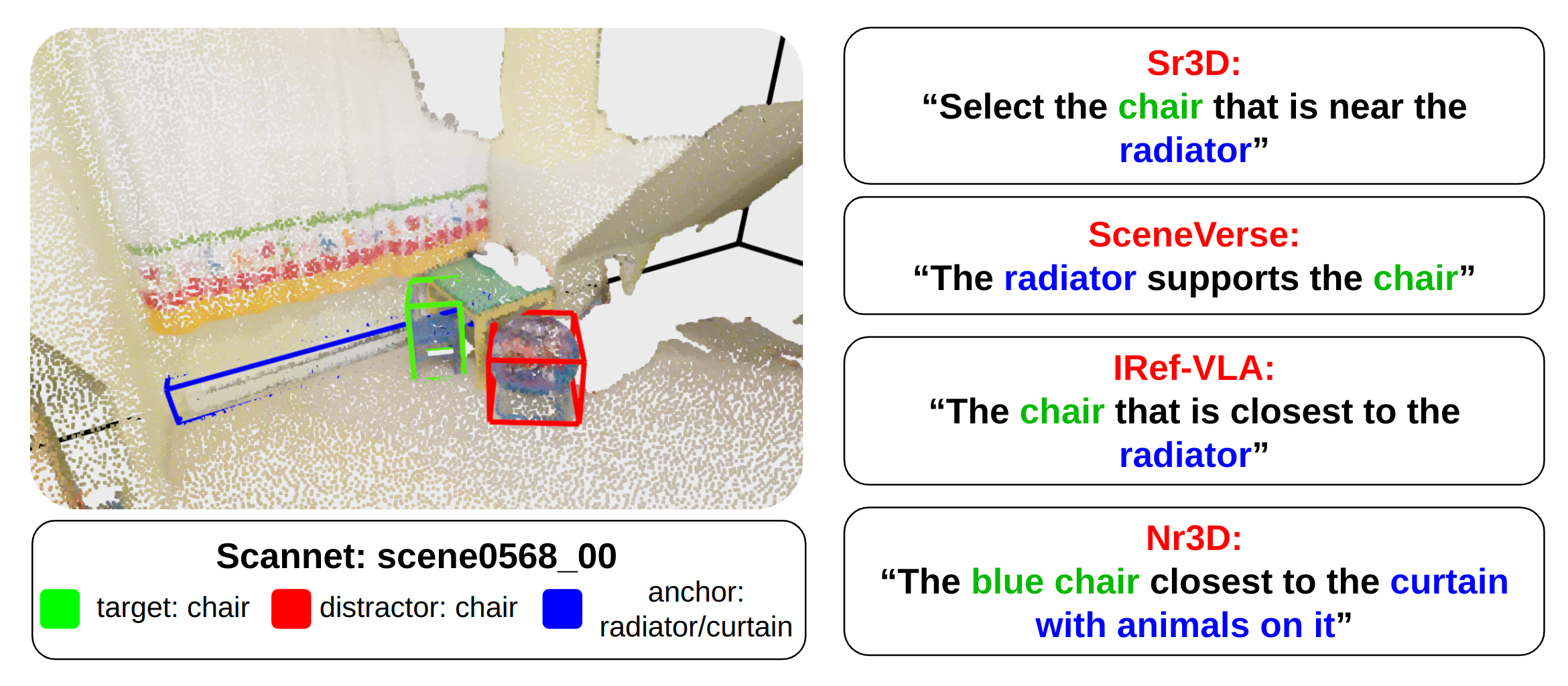}
    \setlength{\belowcaptionskip}{-12pt}
    \caption{A comparison between heuristically generated statements describing a binary spatial relation from Sr3D, Nr3D \cite{achlioptas2020referit3d}, SceneVerse \cite{jia2024sceneverse}, and IRef-VLA. Both chairs are close to the radiator, so using the superlative relation "closest" is the clearest way to disambiguate.}
    \label{fig:statement-compare}
\end{figure}

Referential statements were synthetically generated based on the computed scene graph using a template-based generation method. From the table, synonyms for each relation are used to add variety into the statements. Every statement has at least one semantic relation and only uses object attributes if needed to distinguish the target object. The generated statements are also:
\begin{enumerate}
\item \textbf{View-independent}: The relation predicate for the target object does not depend on the perspective from which the scene is viewed from.
\item \textbf{Unambiguous}: Only one possibility exists in the region for the referred target object.
\item \textbf{Minimal}: Following Grice's maxim of manner \cite{sep-grice}, statements use the least possible descriptors to most clearly disambiguate the target object (Fig. \ref{fig:statement-compare}).
\end{enumerate}

Additionally, the dataset includes "imperfect" referential statements describing non-existent objects. These false statements serve to enhance robustness to noisy language and improve evaluation, as  identifying non-existent objects is a key skill for language grounding. The statements are generated by altering one target or anchor object attribute in existing statements to similar values, ensuring they are contextually similar to true statements.

\section{Baseline Evaluation}
We evaluate our benchmark on both the original referential grounding task and our extended task. First, we compare our data with ReferIt3D \cite{achlioptas2020referit3d} using two SOTA supervised methods for object referential grounding. Then, we implement a graph-search baseline for the task of grounding with imperfect references.

\begin{table}[t!]
\caption{Dataset generalizability on various baseline models}
\label{tab:baselines}
\resizebox{\linewidth}{!}{
\begin{tabular}{@{}llcccc@{}}
\toprule
\multirow{3}{*}{Method} & \multirow{3}{*}{Train Checkpoint} & \multicolumn{4}{c}{Test Set} \\ \cmidrule(l){3-6} 
 & & Sr3D & Nr3D & \multicolumn{2}{c}{IRef-VLA} \\ \cmidrule(l){3-3} \cmidrule(l){4-4}\cmidrule(l){5-6} 
 & & Overall & Overall & ScanNet & Full \\ \midrule
\multirow{4}{*}{MVT \cite{huang2022multi}} & Baseline (Sr3D, reported) & 64.5\% & - & - & - \\ \cmidrule(l){2-6} 
 & Baseline (Sr3D, reproduced) & 59\% & 31.8\% & 29.0\% & 17.2\% \\ \cmidrule(l){2-6} 
 & IRef-VLA-ScanNet & 50.0\% & 29.7\% & 56.0\% & 26.7\% \\ \cmidrule(l){2-6} 
 & IRef-VLA-Full & 41.0\% & 25.9\% & 44.0\% & 47.0\% \\ \midrule
\multirow{5}{*}{3D-VisTA \cite{zhu20233d}} & Baseline (Sr3D, reported) & 76.4\% & - & - & - \\ \cmidrule(l){2-6} 
 & Baseline (Sr3D, reproduced) & 75.7\% & 46\% & 39.2\% & 24.8\% \\ \cmidrule(l){2-6} 
 & SceneVerse \textit{(0-shot text)}  \cite{jia2024sceneverse}  & - & 43.1\% & - & - \\ \cmidrule(l){2-6} 
 & IRef-VLA-ScanNet & 62.4\% & 41.8\% & 63.7\% & 32.3\% \\ \cmidrule(l){2-6} 
 & IRef-VLA-Full & 65.8\% & 44.9\% & 70.8\% & 60.6\% \\ \midrule
\end{tabular}}  
\end{table}

\subsection{Referential Object Grounding }

\subsubsection{Experimental Setup}
To evaluate the effects of scaling up the amount of referential language and number of real-world scenes on the referential grounding task, we train two open-source supervised referential grounding baseline models on our data: MVT \cite{huang2022multi} and 3D-VisTA \cite{zhu20233d}. For the training splits, we use the official ScanNet/ReferIt3D train and validation splits for our ScanNet data, and follow an 80\% train, 20\% validation split for the remaining scenes. To demonstrate generalizability, we test the \textit{zero-shot transfer} capabilities of these models trained on IRef-VLA by training the models first on the ScanNet scenes alone, and then on the full dataset, and evaluating directly on the Sr3D and Nr3D \cite{achlioptas2020referit3d} test sets, which consist of synthetically generated and human-uttered referential statements respectively. %, noting that none of the Nr3D nor Sr3D training language is directly included in training.
Our zero-shot transfer results along with a comparison to the baseline model performance are shown in Table \ref{tab:baselines}. Both models are trained until training loss convergence.

\subsubsection{Generalizability Results}
We observe the following:
\begin{enumerate}[(i)]
    \item Even without seeing any Nr3D statements and without direct training on any Sr3D statements, we observe relatively high accuracies on both test sets when training MVT (50\% on Sr3D, 29.7\% on Nr3D) and 3D-VisTA (62.4\% on Sr3D, 41.8\% on Nr3D) with our ScanNet statements. On the Nr3D test set, we note that the baselines trained on Sr3D perform higher due to similar view-dependent statement distribution, achieving 31.8\% and 46\% accuracy for MVT and 3D-VisTA respectively. However, the small performance differences of 2.1\% with MVT and 4.2\% with 3D-VisTA using the baseline trained on our data shows that our pipeline for synthetically upscaling only the number of referential statements and using new relations without increasing the number of scenes still improves the zero-shot capabilities of object referential models.
    % and is capable of producing natural sounding referential statements.
    \item We observe that increasing the number of training scenes from our dataset further improves the grounding performance of 3D-VisTA on the IRef-VLA ScanNet split from 63.7\% to 70.8\%, on Sr3D from 62.4\% to 65.8\%, and Nr3D from 41.8\% to 44.9\% while MVT instead underfits likely due to being a smaller model. Upscaling the number of our training scenes further improves performance on zero-shot transfer to Nr3D, narrowing the gap for 3D-VisTA between this checkpoint and the Sr3D checkpoint to 1.1\%, despite IRef-VLA containing only view-independent relations. 3D-VisTA trained on our full data also performs $1.8\%$ better than 3D-VisTA trained on the SceneVerse zero-shot text split consisting only of their synthetic statements \cite{jia2024sceneverse}, verifying that better heuristics for generating natural-sounding referential statements improve the effectiveness of upscaling the number of scenes.
    \item Both pre-trained baselines perform poorly generalizing to our IRef-VLA validation sets at 29\%/17.2\% accuracy on IRef-VLA-ScanNet/Full with MVT and 39.2\%/24.8\% with 3D-VisTA, highlighting the difficulty of our benchmark. While there is a significant domain shift with the pre-trained baseline models on our data, those trained on IRef-VLA show a smaller gap when evaluated on Sr3D. This suggests that our dataset's diverse language and scene distribution improves generalization, especially in structured language.
\end{enumerate}

\subsection{Referential Grounding with Imperfect References}

\subsubsection{Experimental Setup}

To establish a quantitative baseline for grounding with imperfect references, we assess methods on two subtasks: 1) identifying existence of objects and 2) suggest alternatives when necessary. We augment SOTA methods for the former and evaluate our graph-search baseline for the latter.
Methods are evaluated on a split of our dataset corresponding to the ReferIt3D \cite{achlioptas2020referit3d} test split. The results can be found in table \ref{tab:classification} and implementation details are further described below.

\begin{table}[H]
\caption{Classification Results for Grounding Object Existence}
\label{tab:classification}
\resizebox{\linewidth}{!}{
\begin{tabular}{@{}l
C{0.12\linewidth}
C{0.12\linewidth}
C{0.12\linewidth}
C{0.12\linewidth}
C{0.12\linewidth}
@{}}
\toprule
Baseline Model & True Positive (TP) & True Negative (TN) & False Positive (FP) & False Negative (FN) & F1-Score \\ \midrule
MVT + Binary Classifier & 66.2\% & 97.1\% & 2.9\% & 33.8\% & 78.3\% \\ \midrule
Graph-Search & 90.4\% & 98.9\% & 1.1\% & 9.6\% & 94.4\% \\ \bottomrule
\end{tabular}}
\end{table}

\begin{figure*}[t!]
\centering
\includegraphics[width=0.9\textwidth]{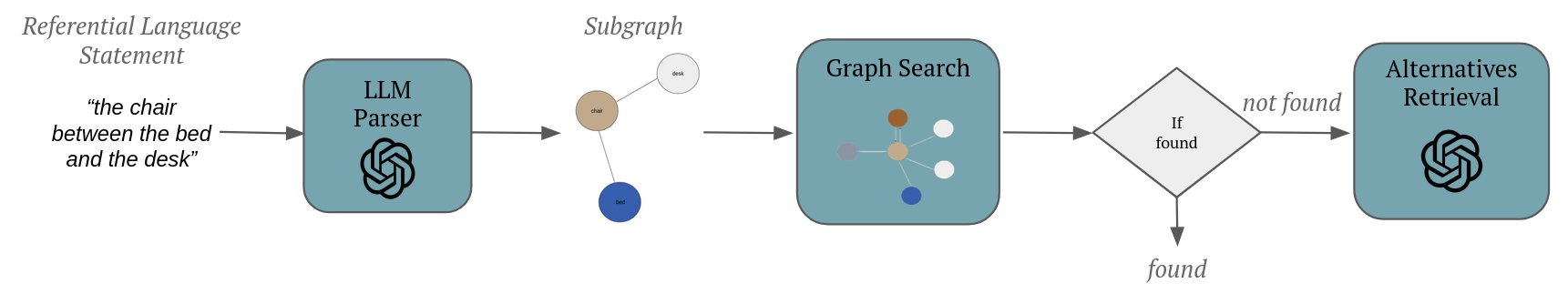}
\caption{Pipeline for graph-search and alternative generation baseline}
\label{fig:graph_search}
\end{figure*}

\textbf{Augmented SOTA Models}:
As existing SOTA referential grounding baselines cannot directly determine whether a referred object is in the scene, an additional binary classification head was added to the MVT model as a point of comparison. The concatenated object features are passed through a simple two-layer MLP and trained with a cross-entropy loss. The additional referential losses are only added if the object truly exists in the scene, ensuring that the object grounding learning is not affected. We use pre-trained checkpoints and finetune with the binary classification loss.

\textbf{Graph Search Baseline}:
To benchmark robustness in grounding with imperfect object references and demonstrate a simple method for alternative generation, we implement a graph-search method using heuristically-generated scene graphs. We first use an LLM, gpt-4o-mini\footnote{https://platform.openai.com/docs/models/gpt-4o-mini}, to parse each statement into a scene subgraph representation with few-shot prompting using five training samples. A given statement is parsed into: target object, anchor objects, attributes, and relation in JSON format, then converted into a subgraph representation where nodes consist of object properties and edges represent relations between objects. We then implement a search method that searches the scene graph for the referenced subgraph. As we are searching for subgraphs and not a single target node, we use breadth-first search to find candidate target nodes, then smaller depth-first searches to find the remaining subgraph. If the exact subgraph is not found, we extract existing referential statements corresponding to partial subgraph matches and prompt the LLM to choose the statement closest to the input statement in a multiple-choice question-answer (MCQA) style prompt. The full pipeline is shown in Fig. \ref{fig:graph_search}.

\subsubsection{Results}
\begin{table}[H]
\caption{Accuracy of Parsing and Alternatives Modules in Graph-Search Baseline}
\label{tab:alternatives}
\resizebox{\linewidth}{!}{%
\begin{tabular}{@{}lcc@{}}
\toprule
Baseline Model & LLM Parsing Accuracy & Average Alternative Similarity \\ \midrule
Graph-Search & 94.0\% & 61\% \\ \bottomrule
\end{tabular}%
} 
\end{table}
We first quantify the LLM parsing accuracy as it directly upper bounds the downstream grounding and alternative scoring. For each statement, we compare the LLM-generated structural output to the ground-truth sub-scene graphs. We achieve a parsing accuracy of 94\% as seen in Table \ref{tab:alternatives}. The results of classifying object existence are in Table \ref{tab:classification}. We note that the graph search baseline is able to find the correct object that exists 90\% of the time (TP rate) using the heuristically-generated scene graphs as the knowledge base, indicating an upper-bound for robust grounding when ground-truth calculated relations are used. In particular, the true negative rate is 97\% for MVT and 98.9\% for the graph-search method, indicating that referential grounding methods can be augmented to explicitly determine when a referred object does not exist as described. When deploying referential grounding methods in the real-world, this would enable robustness of results to changing scenes and mistakes by humans. From Table \ref{tab:alternatives}, scoring LLM-selected alternatives with a simple heuristic results in a score of 61\%, indicating that matching object descriptions alone without direct visual information can set a baseline for alternative selection where over half the aspects match. This can be used as a lower performance bound for comparison to other alternative selection methods developed.

\section{Limitations and Future Work}
As is, our dataset uses synthetically generated language, which, while scalable, lacks view-dependent and allocentric statements common in natural communication. Expanding our dataset to include such statements with possible LLM augmentation or human labeling will enhance the dataset diversity and provide more complex spatial relations. Additionally, the heuristics-based scoring metric for alternatives does not fully capture human preferences or the subtle nuances of alternative suggestions, potentially leading to mismatches with genuine human evaluation criteria. Incorporating human-generated alternatives or having human-scoring of the alternative retrieval method will better capture the subtleties of human intent. Another future direction of work is to explore a multi-turn dialogue setting for specifying navigation goals instead of the single step currently modeled.

\section{Conclusion}
Aiming to advance robust scene understanding for interactive robotic navigation, we introduce IRef-VLA, a novel benchmark dataset for referential grounding with imperfect references. Our benchmark provides a large-scale resource for grounding in 3D scenes while incorporating unique features such as structured scene graphs and imperfect statements to form the novel task of referential grounding with imperfect language. We validate the dataset's diversity and difficulty through baseline experiments with SOTA models, provide a baseline implementation using scene graphs for grounding and alternative generation, and propose metrics to evaluate performance. With this new benchmark and task, we hope to enable the development of generalizable robotics agents robust to imperfections and ambiguities in the real-world when interacting with humans using natural language.


%%%%%%%%%%%%%%%%%%%%%%%%%%%%%%%%%%%%%%%%%%%%%%%%%%%%%%%%%%%%%%%%%%%%%%%%%%%%%%%%

\bibliographystyle{./IEEEtran}
\bibliography{references}

\end{document}